\documentclass[letterpaper, 10 pt, conference]{ieeeconf}  %

\IEEEoverridecommandlockouts                              %

\usepackage{flushend}
\usepackage{graphicx} %
\usepackage{caption}
\usepackage{subcaption}
\usepackage[hidelinks,draft]{hyperref}
\usepackage{amsmath} %
\title{\LARGE \bf
Efficient Gesture Recognition on Spiking Convolutional Networks \\ Through Sensor Fusion of Event-Based and Depth Data 
}

\author{Lea Steffen$^{1}$, Thomas Trapp$^{1}$, Arne Roennau$^{1}$ and R\"udiger Dillmann$^{1}$      
\thanks{$^{1}$All authors are with FZI Research Center for Information Technology,
        76131 Karlsruhe, Germany
        {\tt\small steffen@fzi.de}}%
}

\begin{document}

\maketitle
\thispagestyle{empty}
\pagestyle{empty}

\begin{abstract}
As intelligent systems become increasingly important in our daily lives, new ways of interaction are needed.
Classical user interfaces pose issues for the physically impaired and are partially not practical or convenient.
Gesture recognition is an alternative, but often not reactive enough when conventional cameras are used. 
This work proposes a Spiking Convolutional Neural Network, processing event- and depth data for gesture recognition.
The network is simulated using the open-source neuromorphic computing framework LAVA for offline training and evaluation on an embedded system.
For the evaluation three open source data sets are used. Since these do not represent the applied bi-modality, a new data set with synchronized event- and depth data was recorded.
The results show the viability of temporal encoding on depth information and modality fusion, even on differently encoded data, to be beneficial to network performance and generalization capabilities.

\end{abstract}

\section{INTRODUCTION} \label{sec:introduction}
To allow efficient human-robot collaboration, a meaningful and uncomplicated means of interaction between the human and the machine is necessary.
Historically, interfaces such as buttons, keyboards and joysticks were mainly used here. However, there is now a growing interest in novel interaction schemes.  
Gesture recognition is a versatile way to allow communication despite situational difficulties. This may be due to physical impairments, noisy environments, or simply convenience. \\
Event cameras~\cite{Gallego_survey} are a very interesting technology for this use case, due to their high temporal resolution. 
Hereby, the image acquisition is not transmitted synchronously but events are generated independently for each pixel if a change in illumination exceeds a threshold. 
This technique automatically filters out static objects and creates a sparse representation of the scene. Since only movements are of interest for gesture recognition, this functionality is ideally suited. In addition, there is virtually no motion blur with these sensors, which has great advantages for gesture recognition.
Spiking Neural Networks (SNN), inspired by biological neurons and neural behavior, process temporal information instead of analog signals~\cite{Maass1997, Paugam-Moisy2012}.
SNNs are a subspecies of Artificial Neural Networks (ANN), but they differ greatly in some respects. 
SNNs require significantly less energy and are capable of much faster inference; they also perform particularly well on sparse data. 
However, due to their principle of operation, they are not as accurate as ANN.
A simple neuron model commonly used for SNN is the Leaky Integrate and Fire (LIF)~\cite{Stein1965}.
\begin{figure}[h]
	\centering %
	\includegraphics[width=\linewidth]{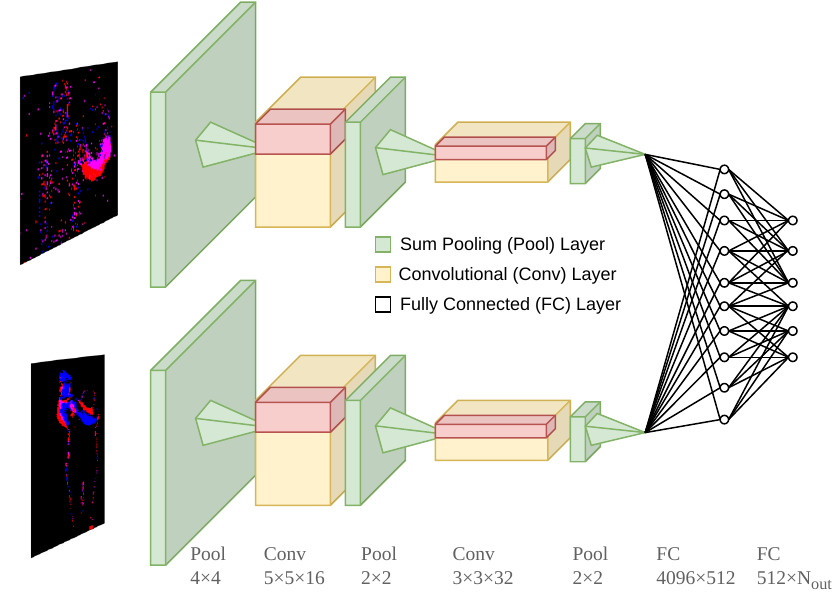}
	\caption{The network architecture includes feature extraction and sensor fusion.
		Event streams and temporally encoded depth data are used as input for the feature extraction. The high-level features are then used together for classification.}
	\label{fig:network_architecture}
\end{figure}
Thereby, the shape of biological action potentials is not considered, instead a uniform event is used. Information is encoded in spike patterns, thus, the time of a spike's appearance carries its essential data content.
An important factor, that makes SNN superior to ANN in terms of performance, is the applied encoding scheme. 
A rate-based encoding for SNN corresponds essentially to the nature of ANNs and leads to the fact that the theoretically possible efficiency increase by SNN cannot be achieved~\cite{Guo2021, Yao2023}. 
Better suited are encoding schemes that take the exact timing of spikes into account, such as time-to-first-spike (TTFS), which enables fast responses within millisecond~\cite{Johansson2004}. TTFS encodes the amplitude as the relative distance in timing between the first spike and a global reference pulse. 
Thereby, shorter relative distances are interpreted as more intense signals, than longer ones.
The majority of deep learning applications rely on gradient descent using the backpropagation of error~\cite{Rumelhart1986}.
However, SNNs are very difficult to train by this method as described in more depth in~\cite{Bengio2015}. 
A central problem is that spikes cannot be differentiated due to the hard threshold of spiking neurons for spike emission. However, this issue was solved with surrogate gradients~\cite{Shrestha2018, Neftci2019}.
A spiking convolutional neural network (SCNN) uses spiking neurons in a convoluted network structure. Thus, a multi-layer SCNN is comprised of alternating convolutional and pooling layers followed by fully-connected layers~\cite{Zhou2019}.  

In this paper, an SCNN is used on temporally encoded depth data and event-based data. For learning, the gradient descent algorithm is used with the help of surrogate gradient functions.
The proposed method is a new direction in gesture recognition using SNNs, not only through the viability of encoding depth data but also through the use of modality fusion to remedy the lack of widely available training data. 

\section{RELATED WORK} \label{sec:related_work}

Methods for gesture recognition on event streams are already presented in~\cite{Lee2012, Amir2017, Chadha2019, Chen2020, Stewart2020}. 
In~\cite{Lee2012} a stereo vision method is applied. The work in~\cite{Amir2017} focuses on the application of the method on neuromorphic hardware. 
Similar to our approach, in~\cite{Chadha2019, Chen2020} a convolutional network architecture is used, however, it contains conventional neurons with no temporal dynamics. 
In~\cite{Stewart2020} gesture recognition is solely a case study for a novel training method, whereby the training process is divided. 
First, in an offline phase, an SNN is trained on GPU using SLAYER and afterwards the network is deployed on a neuromorphic chip, an Intel Loihi. 
Thereby, the last layer is retrained with a surrogate gradient to allow gradient descent for SNN. \\
Sensor fusion for gesture recognition is proposed in~\cite{Wang2019, Ceolini2020, Messikommer2021}.
In~\cite{Ceolini2020} hand gesture recognition is achieved by fusing electromyographic and event data. 
This is realized by a combination of both in a higher-level feature space. In contrast, in~\cite{Wang2019} and~\cite{Messikommer2021} event data is combined with frame-based.\\
Regarding temporal coding, to enable processing of depth data with SNN, previous work is presented in~\cite{Zhou2019}. 
The target application in~\cite{Zhou2019} is object detection on LiDAR point cloud data. 
Thereby, temporal encoding is realized using TTFS which allows sparse processing with a SCNN.

\section{SIGNAL ENCODING} \label{sec:signal_encoding}
\begin{figure*}[t]
	\centering %
	\begin{subfigure}{0.45\textwidth}
		\centering %
		\includegraphics[height=3cm]{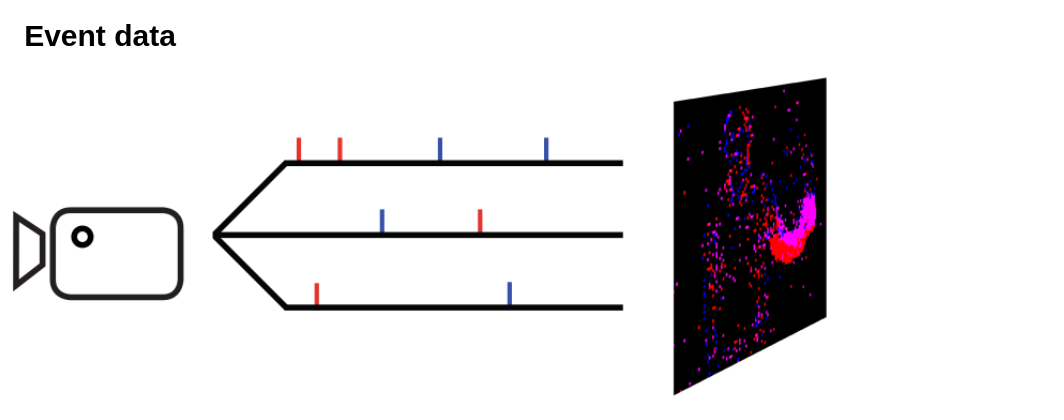}
		\caption{} \label{fig:encoding_event}
	\end{subfigure}\hfil %
	\begin{subfigure}{0.45\textwidth}
		\centering %
		\includegraphics[height=3cm]{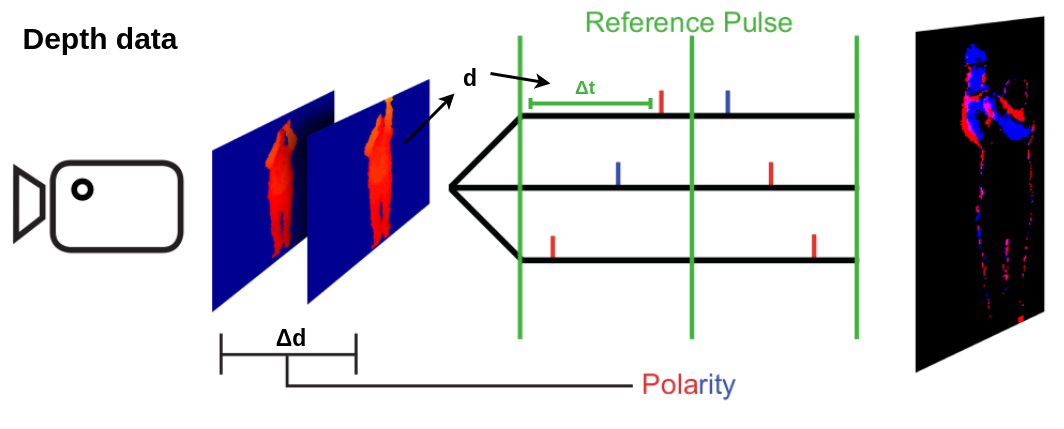}
		\caption{} \label{fig:encoding_depth}
	\end{subfigure}\hfil %
	\caption{Preprocessing is only required for the \subref{fig:encoding_depth} depth data, as the \subref{fig:encoding_event} event stream of the ATIS can be used directly as input for SNN. The depth data are encoded with TTFS to be available in spike trains as well.}
	\label{fig:signal_encoding}
\end{figure*}

The processing is done by an SCNN, as depicted in \autoref{sec:network_architecture}. SNNs work on temporal data, referred to as spike trains. Thus, the sensor signal of both modalities must conform to this format, which is already true for event streams (see \autoref{fig:encoding_event}).
The recorded depth data, however, needs conversion, as visualized in \autoref{fig:encoding_depth}. 
Thus, a pre-processing step is necessary to encode depth data into spike trains. The encoding must contain enough information to allow for classification inference while being fast enough for a real-time setting. Additionally, sparse encoding is very desirable, as it enables efficient processing.
Based on this reasoning, the encoding scheme TTFS is used. The frame rate of the depth sensor is used as a global reference time and a time window is calculated as the inverse of the frames per second (FPS): \(T_s = \frac{1}{f_{fps}}\).
\begin{table}[h]
	\centering
	\small
	\begin{tabular}{c|c}
	
	$t_s$ & time of a specific spike occurrence \\ 
	
	$T_s$ & time window between two spikes \\ 
	
	$I_{in}$ & intensity of the current input \\
	
	$I_{max}$ & maximum intesity across all relevant inputs \\
	
	$d(x,y)$ & the depth value of the pixel (x,y) \\
	
	$d_{max}$ & the maximum depth value of the frame \\
	
	$f_{fps}$ & the frame rate of the depth sensor \\
	
	$\hat{\boldsymbol r}$ & the ground truth rate of spikes \\
	
	$r_{true}$ & the rate of spikes required for prediction \\
	
	$r_{false}$ & base rate of spikes signifying no prediction \\
	
	$\bf{1}$[] & one hot encoded vector \\
	
	$L$ & the loss function \\
	
	\end{tabular}
\caption{Parameter definition for \autoref{eq:depth_encoding}, \ref{eq:target_spike rate} and \ref{eq:spike_rate}.}\label{tab:params}
\end{table}
This allows depth data, which represents the distance to the sensor, to be encoded as a fraction of this time window. Thus, the spike timing is defined relative to a global pulse and either one or no spike is generated in each time window.
The encoding scheme is formalized as
\begin{equation}
t_s =(1 - \frac{I_{in}}{I_{max}}) \ast T_{s} \Rightarrow t_{s} = (1 - \frac{d(x,y)}{d_{max}}) \ast \frac{1}{f_{fps}},
\label{eq:depth_encoding}
\end{equation}
and was developed considering the following aspects (for parameter definition see \autoref{tab:params}):
\begin{enumerate}
	\item \textit{Ordering:} Depth values represent an object's distance to the sensor, with large values representing distant and small values close objects. Encoding this information in a typical TTFS fashion means that distance is conveyed by a temporal delay to the global reference pulse. 
	Thereby an inverse ordering is applied, which represents large distances in shorter spike times than small distances.
	\item \textit{Relativity:} To calculate relative values, either a linear or a logarithmic transform can be used. The linear one directly calculates spike times so the delay grows linearly for the depth value. The direct benefit is its computational simplicity and equal distribution of values along the time frame given by two reference pulses. The logarithmic one, can emphasize differences in depth and result in more differentiated spike timing. However, it leads to clustering of spike times, numerical instabilities and overflow of spike timings into later pulses. Therefore, the linear transform is realized in \autoref{eq:depth_encoding}.
	\item \textit{Polarity:} In SNN synaptic connections are either excitatory or inhibitory~\cite{Maass1997}. This is realized for event cameras by the 1-bit polarity, whereby an ON-event represents an increase in brightness and an OFF-event a decrease~\cite{Gallego_survey}. This dual-channel approach permits easier perception of motion in the scene and allows for more computational possibilities. Depth data does not contain polarity, thus, it was calculated using the differences in pixels between two frames. Thereby, an increase in depth excites the neurons, equivalent to an ON-event, and a decrease inhibits the neurons, equivalent to an OFF-event
\end{enumerate}

\section{NETWORK ARCHITECTURE} \label{sec:network_architecture}
The network topology is a typical CNN structure. To increase the performance the input is first put through a pooling layer to reduce its size. Then the input is further processed through a combination of pooling and convolutional layers. 
The architecture, as visualized in~\autoref{fig:network_architecture} is sub-dived in three parts. The subnets for feature extraction are shown in color and the network for sensor fusion is shown in black.
%
As a first training step, the two network instances for feature extraction are trained on the two different modalities.  
The same network topology is chosen for both networks, the one receiving encoded depth data and the other one taking event data as input.
This enables the option for transfer learning between different modalities, which could further improve the generalization capabilities of the network but is not intended for this approach.
The networks used for this step, contain two spiking convolutional layers which alternate with two spiking pooling layers. This is completed by two spiking fully connected layers for classification. 
The weights of the two networks are used after training to initialize the fusion network, combining the features of the two modalities.
This is done by concatenating data before the final two fully connected layers, allowing the final layers to make use of both features for gesture classification.

\subsection{Training Spiking Networks} \label{sec:training}
As common for classification tasks, the supervised training approach backpropagation, realizing gradient descent, is used.
Regarding the application on SNNs, two challenges need to be addressed~\cite{Bengio2015}:
\begin{enumerate}
	\item the non-differentiable activation function 
	\item the credit assignment problem 
\end{enumerate}
To overcome these issues an approach based on the surrogate gradient methodology~\cite{Shrestha2018} is used. 
Thereby, a LIF model realizes temporal dynamics through a kernel activation function \(\alpha\), which governs the leaky integration dynamics, and the refractory kernel \(\nu\), which governs spike generation and resetting the cell state. This is formalized as:
\begin{equation}
u(t) = \sum_{i=0}^{n} w_i (\alpha \ast s_i)(t) + (\nu \ast s) (t),
\label{eq:lif}
\end{equation}
with \(\ast\) being the convolution operation. A complete parameter definition is provided in \autoref{tab:params}. This formulation as kernel applications allows for the reversal of this dynamic through element-wise correlation. \\
The loss function for the training procedure is formalized by:
\begin{equation}
\hat{\boldsymbol{r}} = r_{true} \mathbf{1}[\text{label}] + r_{false} (1 - \mathbf{1}[\text{label}])
\label{eq:target_spike rate}
\end{equation}
and
\begin{equation}
L = \frac{1}{2}\int_{T_s}({\boldsymbol r}(t) - \hat{\boldsymbol r}(t))^\top {\bf 1}[\text{label}]\,dt.
\label{eq:spike_rate}
\end{equation}
Thereby, \(\mathbf{1}[\text{label}]\) represents a hot encoding of the classification target, meaning a vector where only the index of the corresponding class is \(1\) while all other entries are \(0\).
The target spike rate \(\hat{\boldsymbol{r}}\) is governed by two hyperparameters. Thereby, \(r_{true}\) defines how often spikes should be generated in a time window when the correct class is detected.
In addition, the idea is not to force the network not to spike at all when the wrong class is detected. Instead, the second hyperparameter \(r_{false}\) defines an allowed maximal spike rate for wrong classifications, if this is surpassed it will be penalized by increasing the error. 
The loss is consequently expressed as the distance of the actual spike rate to the target rate.

\section{EXPERIMENTS} \label{sec:experiments}
For the SCNN implementation a spiking-focused Pytorch extension is used, the Lava software framework~\cite{Lava2022}. 
It is an open-source neuromorphic framework, developed and maintained by Intel's Neuromorphic Computing Lab. 
Its goal is to exploit the principles of neural computation while also mapping them to neuromorphic hardware. 
Lava is well suited to implement the learning process, as described in \autoref{sec:training}, because it contains an implementation of the SLAYER Algorithm~\cite{Shrestha2018}.

\subsection{Datasets} \label{sec:datasets}
Several datasets for gesture recognition exist, however, mainly for conventional sensors. Datasets for event cameras are quite rare, even more so for multi-modal data.
\begin{figure*}[t]
	\centering %
	\begin{subfigure}{0.24\textwidth}
		\includegraphics[width=\linewidth]{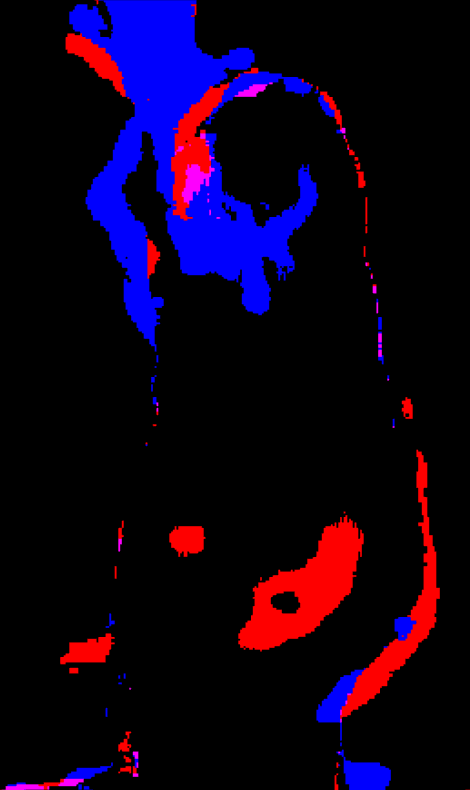}
		\caption{} \label{fig:throw_depth}
	\end{subfigure}\hfil %
	\begin{subfigure}{0.24\textwidth}
		\includegraphics[width=\linewidth]{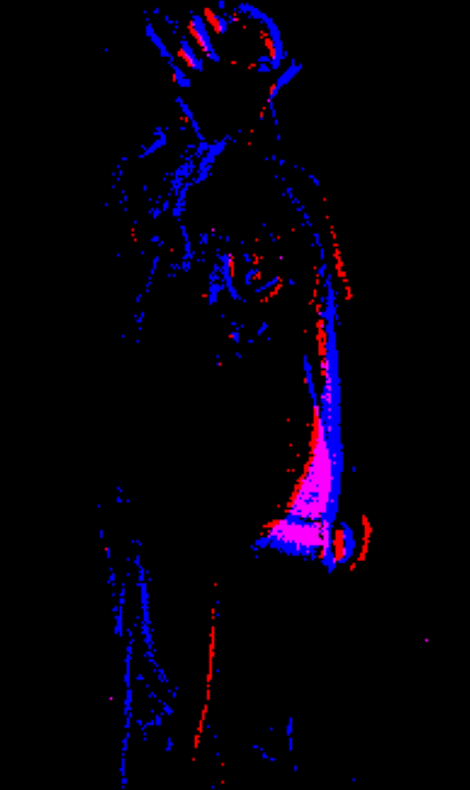}
		\caption{} \label{fig:throw_event}
	\end{subfigure}\hfil %
	\begin{subfigure}{0.24\textwidth}
		\includegraphics[width=\linewidth]{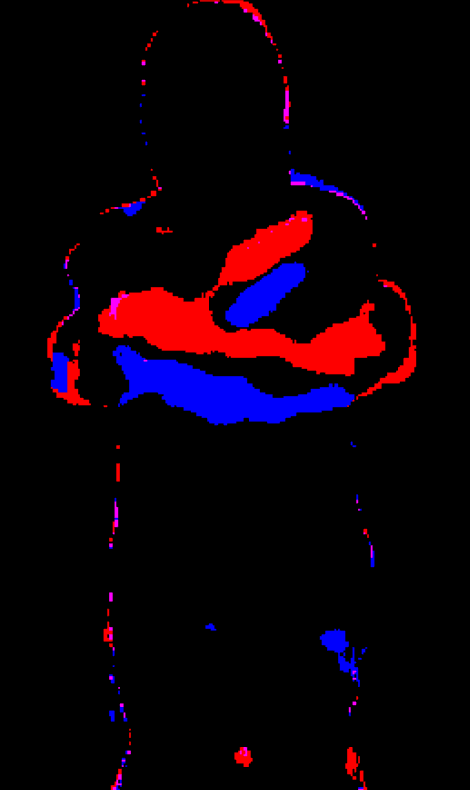}
		\caption{} \label{fig:arms_depth}
	\end{subfigure}
	\begin{subfigure}{0.24\textwidth}
		\includegraphics[width=\linewidth]{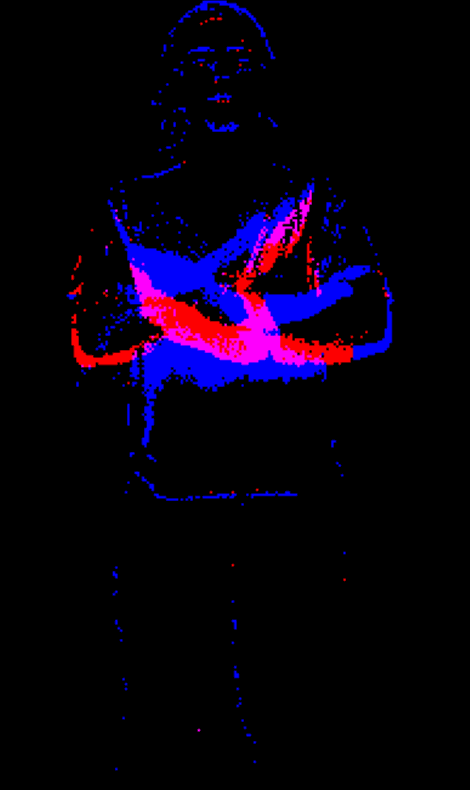}
		\caption{} \label{fig:arms_event}
	\end{subfigure}
	\caption{Examples of gestures from the newly recorded dataset, featuring the encoded depth data from the RealSense in \subref{fig:throw_depth} and \subref{fig:arms_depth}. Respectively, events from the ATIS are visualized in \subref{fig:throw_event} and \subref{fig:arms_event}. Featured
		gestures are throwing an object (\subref{fig:throw_depth}, \subref{fig:throw_event}) and crossing the arms in front of the chest (\subref{fig:arms_depth}, \subref{fig:arms_event}).}
	\label{fig:own_dataset}
\end{figure*}
Unfortunately, no datasets were found featuring both modalities simultaneously. However, several datasets that have at least partial correspondence with the experimental setup and are open source were found. One with event-based data and three with depth data, as shown in \autoref{tab:datasets}.
These datasets diverge quite a bit in size, scope, as well as resolution.
Besides the fact that the datasets listed above do not support any multi-functionality, they also lack consistency between labels and samples are partially unstable and incomplete. Therefore, a new dataset is recorded for evaluation purposes.
\begin{table*}[h!]
	\begin{center}
		\footnotesize
		\begin{tabular}{l|l l l l l }
			
			dataset 	& DVS-Gesture~\cite{DVS_IBM_dataset} & UTD-MHAD~\cite{Chen2015} & UTD-Kinect2~\cite{Chen2016}  & Own \\ \hline
			modality 	& events & depth & depth & event and depth \\ 
			device 		& DVS128 & Kinect & Kinect 2  & ATIS and RealSense \\ 
			datatype 	& aedat3.1 & MAT file & MAT file & ROSbag \\ 
			resolution & \(128 \times 128\) & \(320 \times 240\) & \(512 \times 424\) & \(480 \times 360\) and \(640 \times 480\) \\ 
			\# actions 	& 10 + 1 & 27 & 10  & 30 \\ 
			subjects 	& 29 & 8 & 6  & 2 \\ 
			trials 		& 5 & 4 & 5  & 5 \\ 
		\end{tabular}
	\end{center}
	\caption{A comparison of the datasets used. Thereby, '\# Actions' refers to the number of different classes of gestures in the dataset, 'subjects' represents the number of unique persons featured across the dataset and 'trials' means the amount of sequences per action and subject. Regarding 'resolution', \(480 \times 360\) refers to the event and \(640 \times 480\) to the depth data.}\label{tab:datasets}
\end{table*}
Nevertheless, the above datasets are used for pretraining. Therefore, gestures that are too similar to be effectively differentiated are omitted. 
Furthermore, gestures for which samples exist for both modalities are preferred so that both feature extraction networks are trained on similar data. Strict avoidance of gestures that do not occur in both modalities would limit the set for pretraining to only seven different gestures. \\
\begin{figure}[h]
	\centering %
		\includegraphics[width=0.9\linewidth]{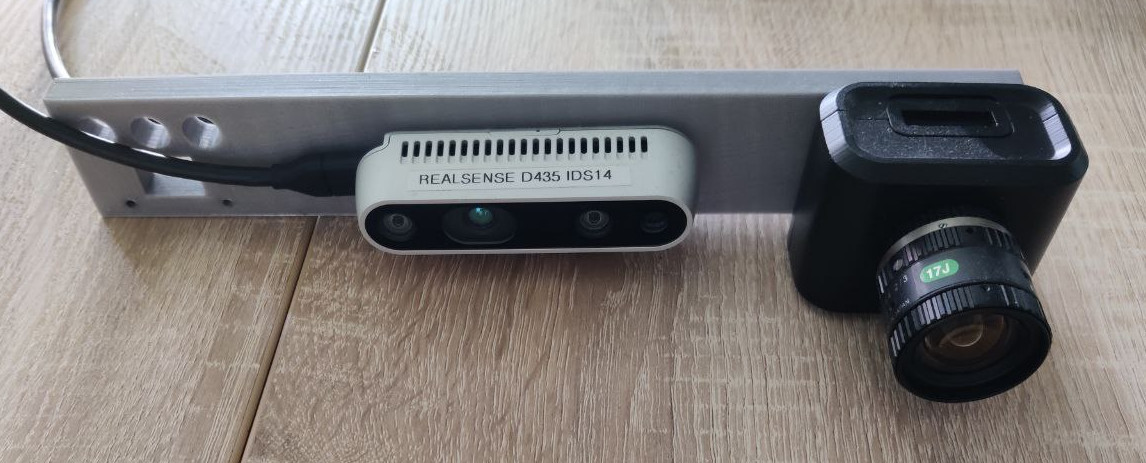}
	\caption{Sensor setup of the ATIS and Intel Realsense, enabling the development of a synchronized bimodal dataset for gesture.}
	\label{fig:sensor_setup}
\end{figure}
The newly developed dataset for this paper combines synchronized recordings of an event camera and a depth sensor. 
The specific sensor model, chosen to provide event data, is the the Prophesee Evaluation Kit Gen3 HVGA-EM, realizing the design of the Asynchronous Time-based Image Sensor (ATIS)~\cite{Posch2011}. 
Respectively, for the depth data, the Intel Realsense is used. 
A large pool of gestures was aimed at when creating the dataset, which can be increased by modifications. 
Nevertheless, the dataset is not overly large, since only two participants made five repetitions each, yielding 300 samples in total.

\subsection{Simulation Parameters} \label{sec:simulation_parameters}
To ease comparisons of networks and keep the network optimization space manageable, neuron parameters and network and simulation parameters are kept constant between runs. 
Thereby, the membrane threshold is set to $1.25$, the voltage decay to $0.3$, the current decay to $1$ and the dropout to $0.1$. 
Additionally, the mean batch norm is used. Regarding simulation parameters, the sequence length is set to $2000~ms$ and the learning rate to $0.05$. 
The epochs are set to $100$ for the single modalities and to $200$ for the fusion network, as less data and more weights need to be handled here. 
As the loss function, the spike rate loss is used.

\section{RESULTS} \label{sec:results}
\begin{figure*}[h]
	\centering %
	\begin{subfigure}{0.48\textwidth}
		\centering %
		\includegraphics[width=\linewidth]{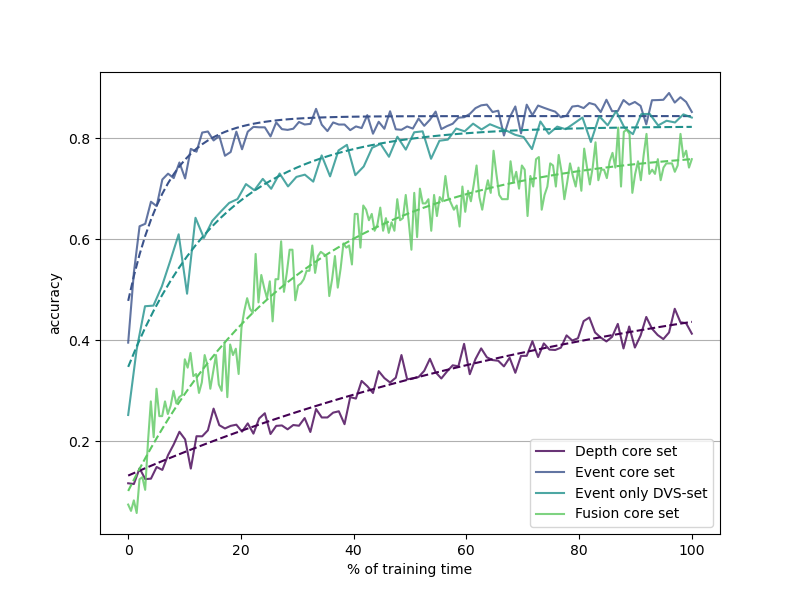}
		\caption{} \label{fig:core_accuracy}
	\end{subfigure}\hfil %
	\begin{subfigure}{0.48\textwidth}
		\centering %
		\includegraphics[width=\linewidth]{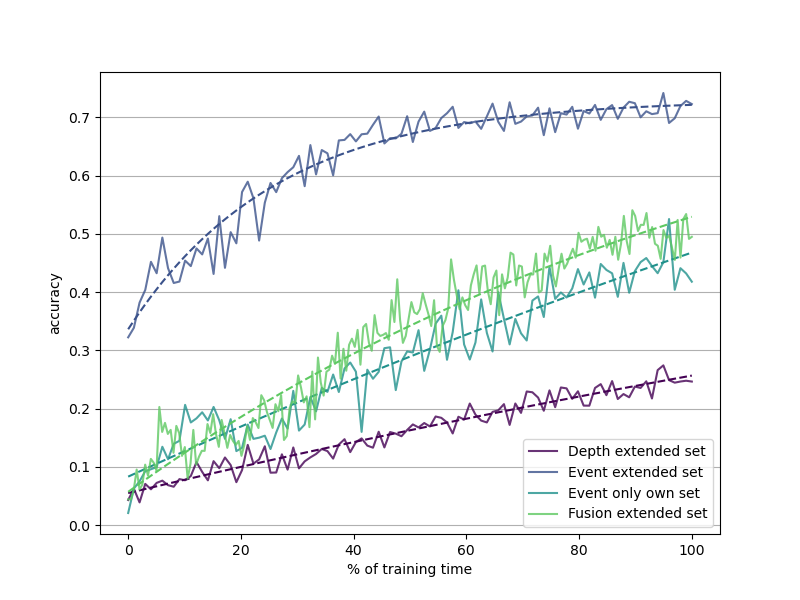}
		\caption{} \label{fig:extended_accuracy}
	\end{subfigure}%
	\caption{Mean accuracies during training between different runs. \subref{fig:core_accuracy} shows the accuracy of networks using the different modalities on a core set of gestures. In \subref{fig:extended_accuracy} the same networks are trained on larger data.}
	\label{fig:dataset_performance}
\end{figure*}
Previous work \cite{Shrestha2018, DVS_IBM_dataset, Stewart2020} suggests that training SNNs on event data for gesture recognition leads to a solid performance. Due to the limited amount of classes that were usually aimed for, a sharp drop in classification accuracy is likely when increasing the number of gestures to be learned. However, regarding depth data used with a temporal encoding scheme, any concrete performance estimates are difficult to obtain.
By combining both modalities, the fusion network is expected to tie the best performance of its sub-networks. Additionally, better generalization capabilities across a larger number of classes are expected. 
The approach is also expected to perform well on small datasets since it can simultaneously make use of information in both modalities. For deployment, the performance in terms of classification strength is expected to be close to the performance in an offline setting. Due to the encoding and processing approach, the network is designed to allow for low-latency applications. \\
\subsection{Offline Training} \label{sec:oflline_training}
The performance during offline training is visualized in \autoref{fig:dataset_performance}. As seen in \autoref{fig:core_accuracy}, the classification of the core set between the different modalities is dominated by the network using event data exclusively, reaching accuracies of up to 88\% on the training and 91.48\% on the testing set. Similar results were achieved by the fusion approach achieving classification accuracies of 86.67\% during training and 80\% during testing. The lowest performance can be observed with the network working on depth data exclusively showing accuracies of 46.79\% and 45.68\% during training and testing respectively. \\
A similar distribution of performance can be observed when training on datasets with more classes. In \autoref{fig:extended_accuracy} it can be seen that the best performance is offered by the network utilizing only event data, although its classification accuracy drops to a maximum of 71.38\% during testing and 79.4\% during training. The classification accuracy using other modalities decreases as well, with the fusion approach reaching up to 54.73\% during training and 58.9\% during testing. Using depth data exclusively results in accuracies up to 26.78\% and 28.09\% during training and testing. \\
To analyze the impact of the choice of dataset on training, a comparison of training the event network on each dataset is carried out. The DVS-Gesture dataset, which only features core labels, was achieved when considered by itself 84.42\% as training and 89.02\% as testing accuracies.
Omission of the DVS-Gesture set, however, shows a significant drop in classification accuracy, only ever reaching a maximum of 64.93\% during training and 43.66\% during testing with both trajectories diverging further. \\
\subsection{Deployment} \label{sec:deployment}
\begin{figure}[h]
	\centering %
	\begin{subfigure}{.4\textwidth}
		\centering %
		\includegraphics[width=\linewidth]{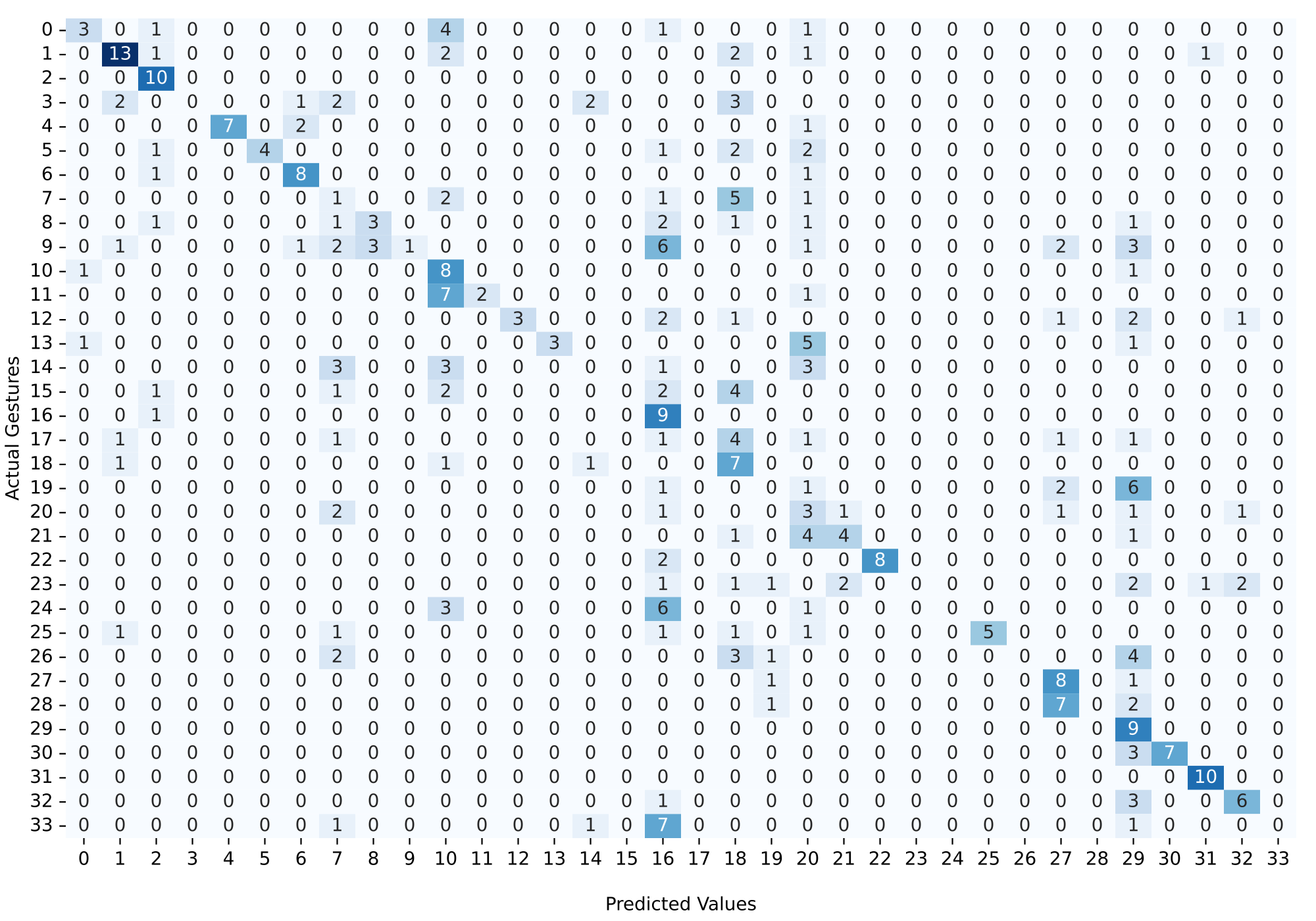}
		\caption{} \label{fig:confusion_lava}
	\end{subfigure}\hfil %
	\begin{subfigure}{.08\textwidth}
		\centering %
		\includegraphics[width=\linewidth]{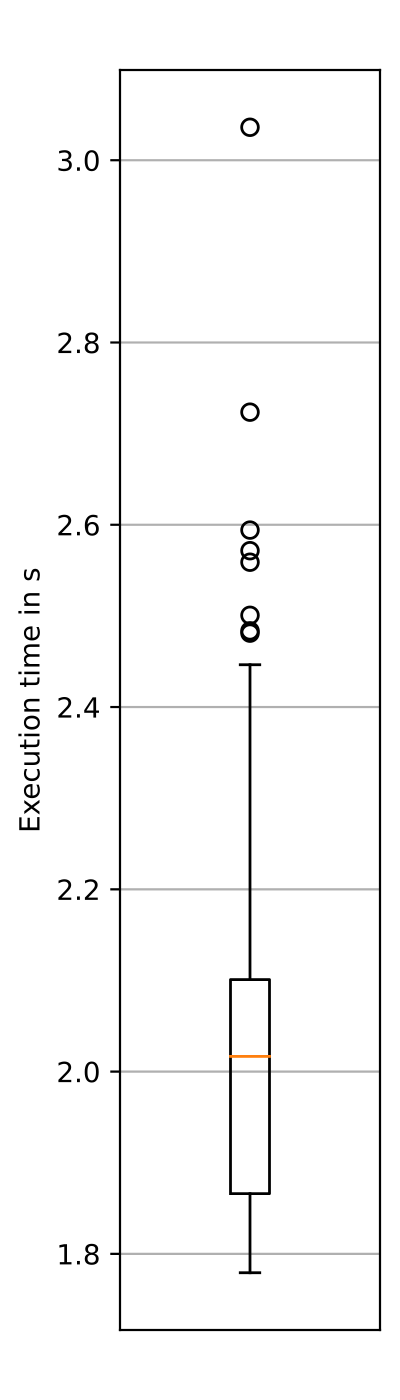}
		\caption{} \label{fig:exec_times_lava}
	\end{subfigure}%
	\caption{Experimental results of the network deployment onto an NVIDIA Jetson Xavier AGX. \subref{fig:confusion_lava} shows the confusion matrix of the classification output, while \subref{fig:exec_times_lava} shows the distribution of execution times per sample of 2000ms length}
	\label{fig:deployment_eval}
\end{figure}
The results when deploying the network onto embedded hardware are summarized in \autoref{fig:deployment_eval}. The accuracy concerning the complete dataset is 39.55\%, which is significantly lower than the expected mean accuracy of 56.81\% (mean of training and testing performance). In the confusion matrix, in \autoref{fig:confusion_lava}, a distinguishable diagonal can be seen alongside some prediction outliers which tend to accumulate into rows of increased activity. The execution time of the processing of a 2-second sample of spikes is visualized in \autoref{fig:exec_times_lava}, ranging from 1.78s to 3.04s with a mean execution time of 2.02s. 
 
 \subsection{Discussion} \label{sec:discussion}
The network performance using only event data shows performance analogous to related work and serves as a baseline performance test for the own dataset. While the depth and fusion approaches are underwhelming in terms of raw classification accuracy, it does seem that at least some learning can be achieved using only depth data. The approach using modality fusion also stands out for its ability to make use of very limited amounts of training data at the cost of needing multi-modal recordings. This is amplified by its solid performance on a set on which a purely event-based approach overfitted. \\
One potential source of the depth networks' low performance is assumed in the lack of hyperparameter analysis. While a lot of different configurations and parameterizations of the depth data encoding, neuron models and network topologies have been tested, the testing was far from exhaustive. This could also explain the unexpected drop in performance of the fusion approach when compared to a single modality network.
Another aspect to consider is the heterogeneity of the datasets. While previous event-based approaches mainly relied on a single dataset with a consistent setup, the addition and omission of different datasets under different conditions may impact performance negatively, as can be seen with the training runs using single sets. \\
The results of the deployment show lower classification accuracies than in the offline settings and execution times that are on average slightly higher than the chosen sample length. These circumstances prevent the system from being run in a real-time manner in its current state. However, optimization might decrease execution times further. Its current state already manages to show the viability of limited power simulation of SNNs on embedded hardware.
More specifically the confusion matrix seems to indicate an accumulation of activity and therefore output spikes even for unrelated classes. Most of the wrong predictions were not a case of sensibly similar activities that have movements in common, but rather specific neurons which show a high frequency of activation (e.g. Neurons 10, 16, 18, 29 in \autoref{fig:confusion_lava}). This could indicate variability in the model execution due to underlying architectural differences. \\
In summary, while this approach of depth encoding and multi-modal fusion does not yield a new benchmark for classification accuracy on gesture recognition or performance in terms of processing speed and resource management, it does show that temporal encoding of depth data is useful for gesture recognition tasks. Combined with multi-modal approaches effective use of limited datasets for training can be achieved.

\section{CONCLUSIONS AND FUTURE WORKS} \label{sec:conclusion}
In this work, the temporal encoding of depth images as well as their fusion with event data was conceptualized and implemented. This approach was then used to train and compare networks for exclusive event- and depth-based gesture recognition, as well as a network using sensor fusion of both modalities. To this end, a multi-modal dataset was recorded using an event and depth camera stereo setup. The trained network was then deployed and evaluated on an embedded system.
Thereby, the viability of using temporally encoded depth data in gesture recognition was shown. Additionally, fusing the modalities proved useful in remedying the typical signs of overfitting on the small available dataset. This indicates good generalization capabilities and effective use of the samples. Finally, the deployment of SNN models onto embedded systems featuring conventional hardware was shown to be possible.
These achievements show the potential benefits of using encoded modalities in SNNs, which allow for the integration of readily available data for the training of SNNs. In particular, the approach for sensor fusion enables a wide range of potential new research fields without the strict need for excessively large available datasets. The possibility of SNNs running on embedded systems further allows for contesting conventional deep learning models in many applications.
To compete with the classic deep learning applications, further optimization of the deployment workflow is needed. An in-depth analysis of the encoding techniques regarding depth data and sensor fusion may also lead to more competitive performance in SNNs.
In a broader context, the reactive gesture recognition approach presented here can be used in a shared workspace between humans and robots. A neural system as presented in~\cite{Steffen2022_iros}, where highly reactive path planning is performed in the configuration space by mimicking neural structures would be particularly interesting for this.

\section*{ACKNOWLEDGMENT}
This research has been supported by the European Union's Horizon 2020 Framework Programme for Research and Innovation under the Specific Grant Agreement No. 945539 (Human Brain Project SGA3).

\bibliographystyle{IEEEtran}
\bibliography{bib}

\end{document}